\pdfoutput=1

\documentclass[11pt]{article}

\usepackage[]{acl}

\usepackage{times}
\usepackage{latexsym}

\usepackage[T1]{fontenc}

\usepackage[utf8]{inputenc}
\usepackage{graphics}

\usepackage{microtype}

\usepackage{microtype}
\usepackage{graphicx}
\usepackage{subfigure}
\usepackage{booktabs} %

\usepackage{makecell}

\usepackage{amsmath,amsfonts,bm}

\def\eqref#1{equation~\ref{#1}}

\def\1{\bm{1}}

\DeclareMathAlphabet{\mathsfit}{\encodingdefault}{\sfdefault}{m}{sl}
\SetMathAlphabet{\mathsfit}{bold}{\encodingdefault}{\sfdefault}{bx}{n}

\usepackage{CJKutf8}

\usepackage{hyperref}
\usepackage{url}
\usepackage{graphicx}
\usepackage{booktabs}
\usepackage{multirow, multicol}
\usepackage{diagbox}

\usepackage{tabularx}
\usepackage{xspace}
\usepackage{amsmath}
\usepackage{float}
\usepackage{xcolor}
\usepackage{tablefootnote}

\usepackage{pifont}
\usepackage{amssymb}

\usepackage{enumitem}

\newcommand{\chin}[1]{\begin{CJK*}{UTF8}{gbsn}#1\end{CJK*}}

\newcommand{\kor}[1]{\begin{CJK}{UTF8}{mj}#1\end{CJK}}

\newcommand{\xwoz}[0]{X-RiSAWOZ\xspace}

\newcommand{\dialogue}[0]{Common Dialogue\xspace}

\DeclareSymbolFont{extraup}{U}{zavm}{m}{n}
\DeclareMathSymbol{\varheart}{\mathalpha}{extraup}{86}
\DeclareMathSymbol{\vardiamond}{\mathalpha}{extraup}{87}

\newcommand\blfootnote[1]{%
  \begingroup
  \renewcommand\thefootnote{}\footnote{#1}%
  \addtocounter{footnote}{-1}%
  \endgroup
}

\title{X-RiSAWOZ: High-Quality End-to-End Multilingual Dialogue Datasets \\ and Few-shot Agents}

\author{
\fontsize{11pt}{11pt}\selectfont
 \makecell{
 $^\clubsuit$Mehrad Moradshahi$^1$ \; $^\clubsuit$Tianhao Shen$^2$ \;  Kalika Bali$^3$ \; Monojit Choudhury$^3$ \\ Gaël de Chalendar$^4$ \; Anmol Goel$^5$ \; Sungkyun Kim$^6$ \; Prashant Kodali$^5$ \\ Ponnurangam Kumaraguru$^5$ \; Nasredine Semmar$^4$  \; Sina J. Semnani$^1$ \; Jiwon Seo$^6$ \\ Vivek Seshadri$^{3,7}$ \; Manish Shrivastava$^5$ \; Michael Sun$^1$ \;  Aditya Yadavalli$^7$ \\ Chaobin You$^2$ \; $^\vardiamond$Deyi Xiong$^2$ \; $^\vardiamond$Monica S. Lam$^1$  
 }
 \vspace{2pt}
\\
\fontsize{7.3pt}{7.3pt}\selectfont
\makecell{
$^1$ Computer Science Department, Stanford University, Stanford, USA
$^2$ College of Intelligence and Computing, Tianjin University, Tianjin, China \\
$^3$ Microsoft, Bangalore, India 
$^4$ Université Paris-Saclay, CEA, List, Palaiseau, France \\
$^5$ International Institute of Information Technology, Hyderabad, India \\ 
$^6$ Computer Science Department, Hanyang University, Seoul, South Korea 
$^7$ Karya Inc., India  
}
}

\begin{document}
\maketitle

\begin{abstract}
Task-oriented dialogue research has mainly focused on a few popular languages like English and Chinese, due to the high dataset creation cost for a new language. 
To reduce the cost, we apply manual editing to automatically translated data. We create a new multilingual benchmark, \xwoz, by translating the Chinese RiSAWOZ to 4 languages: English, French, Hindi, Korean; and a code-mixed English-Hindi language.
\xwoz has more than 18,000 human-verified dialogue utterances for each language, and unlike most multilingual prior work, is an end-to-end dataset for building fully-functioning agents. 

The many difficulties we encountered in creating \xwoz led us to develop a toolset to accelerate the post-editing of a new language dataset after translation. This toolset improves machine translation with a hybrid entity alignment technique that combines neural with dictionary-based methods, along with many automated and semi-automated validation checks. 

We establish strong baselines for \xwoz by training dialogue agents in the zero- and few-shot settings where limited gold data is available in the target language. Our results suggest that our translation and post-editing methodology and toolset can be used to create new high-quality multilingual dialogue agents cost-effectively. Our dataset, code, and toolkit are released open-source.\footnote{\url{https://github.com/stanford-oval/dialogues}}
\blfootnote{$\clubsuit$ Co-first authors $\vardiamond$ Co-corresponding authors}

\end{abstract}
\section{Introduction}
\label{intro}

In recent years, tremendous effort has been put into the research and development of task-oriented dialogue agents; yet, it has been mainly focused on only a handful of popular languages, hindering the adoption of dialogue technology around the globe. Collecting dialogue data from scratch for a new language is ideal but prohibitively expensive and time-consuming, leading to the current lack of reliable multilingual dialogue benchmarks. 

In recent years, several non-English task-oriented dialogue (ToD) datasets have been created. These datasets are either collected from scratch ~\cite{risawoz,zhu-etal-2020-crosswoz}, synthesized using a state machine with manually written templates, and paraphrased for fluency by crowdworkers ~\cite{lin2021bitod}, or manually translated from another language ~\cite{li2021multi-domain}. All of these approaches are labor-intensive, costly, and time-consuming; such investment is unlikely to be made for less widely spoken languages.

This motivates the development of zero and few-shot techniques that can produce a usable agent in a new language with no or only a few gold training dialogues in the target language.
Concurrent with this work, ~\citet{ding-etal-2022-globalwoz,zuo2021allwoz,hung2022multi2woz} adopt a \emph{ translation and manual post-editing} process where data is translated with neural machine translation models first, and then post-edited by crowdworkers. This approach has shown promise on MultiWOZ; however, reported zero- and few-shot accuracies show a big degradation in performance compared to full-shot accuracy in the source language. Besides, the performance of the agent in the original language was not good to begin with, in part due to misannotations in the dataset~\cite{eric2019multiwoz}. Lastly, these datasets either focus only on the subtask of Dialogue State Tracking (DST)~\cite{ding-etal-2022-globalwoz} or auxiliary tasks such as Response Retrieval~\cite{hung2022multi2woz}, or are too small~\cite{zuo2021allwoz} to train end-to-end dialogue agents that require policy, interactions with databases, and response generation components.

Our overall goal is to make task-oriented dialogue research in major languages available to low-resource languages. The key is to produce high-quality few-shot training, validation, and test set with as little manual effort as possible to enable zero-shot or few-shot training. We describe below our contributions towards this goal. 

\subsection{Data Translation Techniques and Toolset}

Machine translation followed by human post-editing has been used as a method for extending monolingual NLP datasets to new languages~\cite{yang2019paws,ziemski2016united,giannakopoulos2011tac,conneau2018xnli}. However, we discovered human post-editing to be the main pain point in creating new dialogue datasets. The process is costly, requiring a lot of back-and-forth among developers, translators, and annotators. Even after several rounds, the results are still not adequate. To alleviate this, we devised a scalable methodology and an associated toolkit that automates parts of this process, and aids translators and annotators to iteratively check their work themselves without developer supervision. This allows fast and accurate creation of a new dialogue dataset annotated with slot values for a new language.

We show that the entity-aware translation technique proposed by~\citet{moradshahi-etal-2023-zero} is also applicable to other end-to-end dialogue datasets. We combine this technique with a dictionary-based alignment where multiple translations are generated for each entity individually (i.e. without context), using the same translation model used to translate the sentence. Then, the translated sentence is scanned to match any of the translation candidates, resulting in an improvement in the agent's performance.

Furthermore, we automatically check each step of data translation to ensure annotation consistency between dialogue utterances and API calls to the database. We are releasing this toolkit open-source for reproducibility as well as a resource for others.

\begin{table}\fontsize{8}{10}\selectfont
\centering
\newcolumntype{C}{>{\centering\arraybackslash}X}
\newcolumntype{L}{>{\arraybackslash}m{3.5cm}}
\begin{tabularx}{\linewidth}{l|C|C|C}
\toprule
\multirow{3}{*}{} & \multicolumn{3}{L}{\bf Dataset}  \\
\hline
\bf{} & {\bf Few-shot} & {\bf Validation} & {\bf  Test}\\
\hline
\# Domains & 12 & 12 & 12  \\ 
\# Dialogues & 100 & 600 & 600 \\
\# Utterances & 1,318 & 8,116 & 9,286 \\
\# Slots   & 140 & 148 & 148 \\
\# Values & 658 & 2,358 & 3,571 \\
\bottomrule
\end{tabularx}
\vspace{-1em}
\caption{Statistics for the few-shot, validation, and test.}
\label{tab:data_stats}
\vspace{-2.2em}
\end{table}

\subsection {X-RiSAWOZ Dataset} 
We created X-RiSAWOZ, a multi-domain, large-scale, and high-quality task-oriented dialogue benchmark, produced by translating the Chinese RiSAWOZ data to four diverse languages: English, French, Hindi, and Korean; and one code-mixed English-Hindi language. X-RiSAWOZ is an improvement over previous works in several aspects:
\begin{itemize}[leftmargin=*,topsep=0pt,noitemsep]
\item
\textbf{End-to-End}: Contains translations for all parts of dialogue including user and agent utterances, dialogue state, agent dialogue acts, and database results. 

\item
\textbf{Larger}: RiSAWOZ is larger than MultiWOZ and covers a total of 11,200 dialogues with 151,982 turns. It also covers 12 domains compared to 7.
In addition to translating validation and test data, we also sample 100 dialogue examples from the training set and translate them using the same process to use as few-shot training data. This way, \xwoz can be used to experiment with few-shot techniques as well as zero-shot.

\item
\textbf{Higher Quality}: We choose RiSAWOZ as it exhibits the lowest misannotation rate among popular dialogue benchmarks as shown by ~\citet{moradshahi2021contextual}. The data translation methodology described above reduces the mismatch between entities in the sentence and annotations, meaning that our translation process does not introduce new misannotations.

\item 
\textbf{Cheaper}: First, the methodology and toolset reduce the amount of post-editing effort needed. Second, instead of using commercial translation systems such as Google Translate, we rely on open-source multilingual translation models such as MBART~\cite{liu2020multilingual} for the translation of training data. This reduces the translation cost by at least 100x which could otherwise be a prohibiting factor when building datasets for new languages. 

\end{itemize}

\subsection{Experimental Results}

We establish strong baseline results for our new \xwoz dataset. In the full-shot setting, our model produces a new SOTA on the original Chinese dataset. With few-shot training, across languages, our model achieves between 60.7-84.6\% accuracy for Dialogue State Tracking (DST), 38.0-70.5\% accuracy for Dialogue Act (DA), and 28.5-46.4\% for BLEU score when evaluated using gold data as the conversational context. Cumulatively over a conversation, our model achieves 17.2\%, 11.9\%, 11.3\%, 10.6\%, and 2.3\% on Dialogue Success Rate (DSR), respectively. The remaining gap between zero or few-shot results on new languages and the full-shot results on Chinese creates opportunities for research and finding new techniques to further improve the dialogue agent performance.

\section{Related Work}
\label{related}

\subsection{Multilingual Dialogue Datasets}

MultiWOZ~\cite{multiwoz1, multiwoz2, multiwoz21}, CrossWOZ~\cite{zhu-etal-2020-crosswoz}, and RiSAWOZ~\cite{risawoz} are three monolingual Wizard-Of-Oz multi-domain dialogue datasets for travel dialogue agents.
For the 9th Dialog System Technology Challenge (DSTC-9)~\cite{dstc9}, MultiWOZ was translated to Chinese and CrossWOZ was translated to English using Google Translate. A portion of their evaluation and test sets were post-edited by humans, while the training set remained entirely machine translated. \citet{moradshahi2021contextual} translated RiSAWOZ to English and German using open-source machine translation models with alignment. However, the validation and test data were not verified by humans, resulting in potentially over-estimating the accuracy of agents.
Several works~\cite{ding-etal-2022-globalwoz,zuo2021allwoz,hung-etal-2022-multi2woz} continued translation of MultiWOZ to other languages. For example, GlobalWOZ translates to several languages, with human translators post-editing machine-translated dialogue templates, and filling them with newly collected local entities. However, these works address only one or two subtasks of a full dialogue, and therefore training an end-to-end agent is not possible with them.  

Different from these translation-based approaches, \citet{lin2021bitod} introduced BiToD, the first bilingual dataset for \emph{end-to-end} ToD modeling. BiToD uses a dialogue simulator to generate dialogues in English and Chinese, and asks crowdworkers to paraphrase them for naturalness. This simulation-based approach eliminates the need for translation but requires hand-engineered templates and savvy developers with knowledge of the target language and dialogue systems. Besides, paraphrasing the entire dataset is costly.

\subsection{Cross-Lingual Approaches for ToD}

With the advent of pre-trained language models, contextual embeddings obtained from pre-trained multilingual language models~\cite{devlin2018bert, mt5, liu2020multilingual} have been used to enable cross-lingual transfer in many natural language tasks, including task-oriented dialogue agents. Unfortunately, most of this work has only focused on the DST subtask, which is a limitation we aim to rectify with this paper. 

To further improve the cross-linguality of these embeddings, \citet{tangetal-codeswitch-2020} and \citet{moghe-etal-2021-cross} proposed fine-tuning multilingual BERT on a synthetic code-switching dataset.
\citet{glavas-etal-2020-xhate} performed language adaptation by using intermediate masked language modeling in the target languages and improving zero-shot cross-lingual transfer for hate speech detection task.

Using machine translation for multilingual dialogue tasks has also been studied. ~\citet{uhrig-etal-2021-translate} used machine translation during inference to translate to English for semantic parsing. Instead, ~\citet{bootstrap} use machine translation to generate semantic parsing data to train a semantic parser in the target language which leads to better results. ~\citet{moradshahi-etal-2023-zero,nicosia2021translate} proposed using alignment to improve the quality of translated data by ensuring entities are translated faithfully.

\section{The End-to-End ToD Task}
\label{task}
In end-to-end task-oriented dialogues, a user speaks freely with an agent over several turns to accomplish their goal according to their intents (e.g., "book a hotel with at least 5 stars"). In each turn, the agent must access its database if necessary to find the requested information (e.g., find a hotel that meets user constraints), decide on an action (e.g., present the information to the user or ask for additional information), and finally respond to the user in natural language based on the action it chooses. Following ~\cite{moradshahi-etal-2023-zero}, we decompose a dialogue agent into four subtasks: 
\begin{enumerate}[leftmargin=*,topsep=0pt,noitemsep]
\item
\emph{Dialogue State Tracking (DST)}: Generate the new belief state, for the current turn based on the previous belief state, the last two agent dialogue acts, and the current user utterance.
\item
\emph{API Call Detection (ACD)}: Determine if an API call is necessary to query the database.
\item
\emph{Dialogue Act Generation (DAG)}: Generate  the agent dialogue act based on the current belief state, the last two agent dialogue acts, the user utterance, and the result from the API call.
\item
\emph{Response Generation (RG)}: Convert the agent dialogue act to produce the new agent utterance.
\end{enumerate}

\section{The \dialogue Interface}
\label{sec:interface}

Over the years, various ToD datasets have been introduced~\cite{multiwoz1,byrne2019taskmaster,zhu2020crosswoz,risawoz,lin2021bitod}, each with its own representation, making it difficult for researchers to experiment with different datasets. To facilitate experimentation, we have developed \dialogue, a standard interface for ToD tasks. This interface defines a unified format for datasets, their annotations, ontologies, and API interfaces. We show that the most widely-used recent dialogue datasets (such as MultiWoZ, RiSAWOZ, and BiToD) can be converted to this representation with a simple script. The standardization lets all different datasets be processed with the same software and models, significantly reducing the implementation time and cost.

Previously, other libraries such as ParlAI~\cite{miller2017parlai}, ConvLab~\cite{zhu2020convlab2,zhu2022convlab3}, and Nemo~\cite{kuchaiev2019nemo} were introduced so researchers can work with different dialogue datasets and interact with the trained models. However, these libraries are limited. They either do not provide a standard abstraction, making it difficult to add new datasets, or a modular interface that can connect with other code bases, requiring new models to be implemented in their repository before they can be used. Additionally, the training code needs to be modified to support a new dataset or language for an existing dataset.

\begin{figure*}
    \centering
    \includegraphics[width=0.85\textwidth]{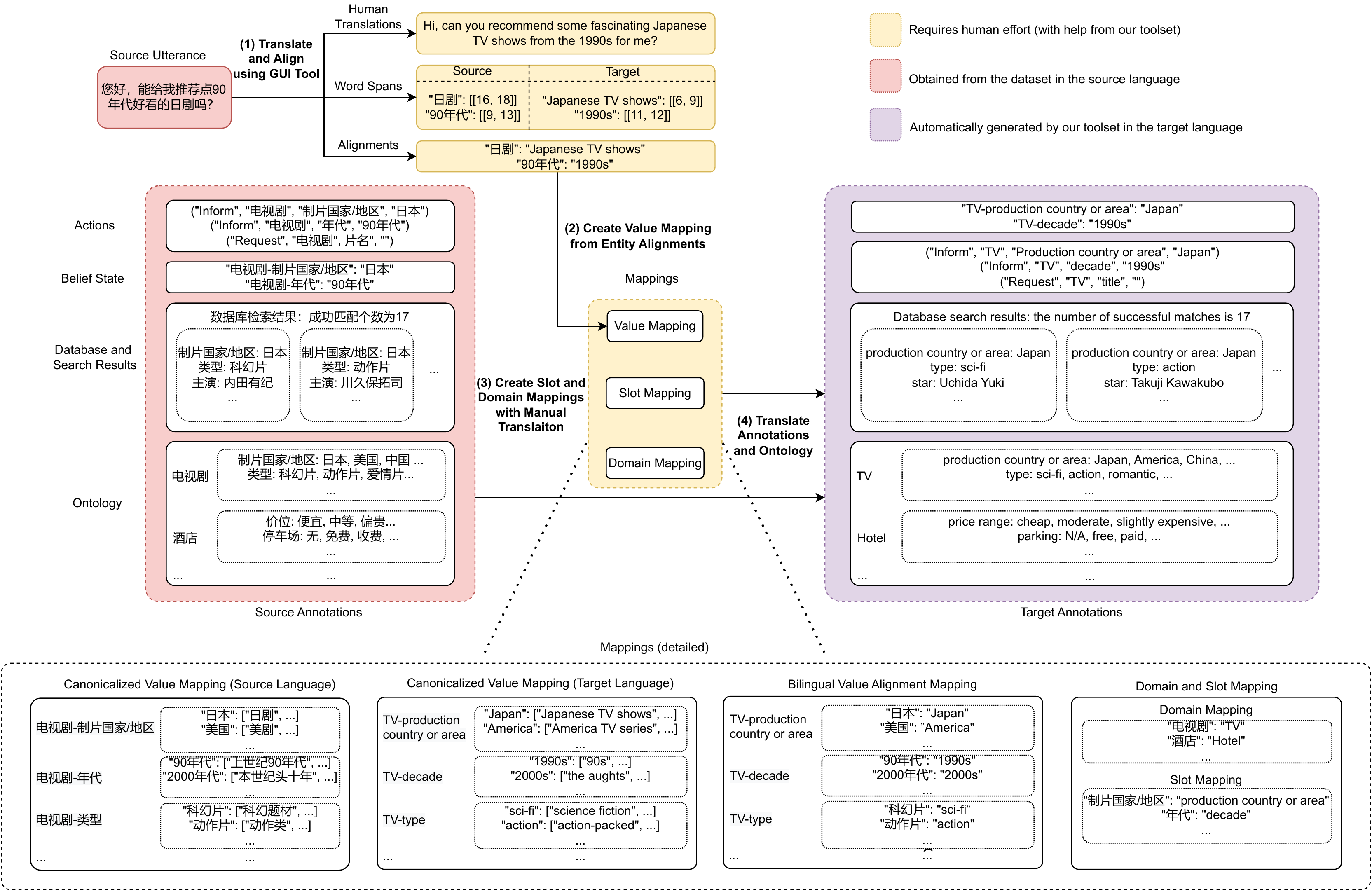}
    \vspace{-0.8em}
    \caption{The translation and annotation process of X-RiSAWOZ from Chinese to English. There are 4 major steps: (1) Translate utterances and provide entity alignments between source and target sentences using the UI tool. (2) Create the value mapping using entity alignments. (3) Create slot and domain mappings by manually translating them from Chinese. (4) Translate slot values in the annotations and ontology using the value mapping.}
   \vspace{-0.8em}
    \label{fig:dataset-translation-process}
\end{figure*}
\section{Dataset Creation}
\label{dataset}

In this section, we describe the process used to extend RiSAWOZ to the new languages. The original RiSAWOZ dataset is in Chinese. We manually translate the validation data (600 dialogues), test data (600 dialogues), and 1\% of the training dataset (100 dialogues), which we refer to as {\em few-shot}, from Chinese to English. For other languages, we use English as the source language, since bilingual speakers of English and the target language are more accessible than Chinese and the target language. Since the English data is manually translated, this approach avoids double translationese~\citep{vanmassenhove2021machine} and ensures the best data quality. We machine-translate the English data and manually post-edit the translation for fluency and correctness.
Besides the few-shot data, we also machine-translate all of the Chinese training data into each of the languages (including English) and train with them; we refer to training with just this data set as {\em zero-shot}, since no human labor is used during  dataset creation.

In the following, we discuss the steps and methods for preparing data for translation, including building alignment between entities and performing iterative quality checks. We also describe how to create the target language ontology, which serves as a database for API calling and provides a mapping between source and target language entities.

\subsection{Translation and Alignment for Few-Shot, Validation, and Test Data}
\subsubsection{From Chinese to English}
Figure~\ref{fig:dataset-translation-process} shows the process used to translate the Chinese dataset to English.  
First, human professional translators manually translate the Chinese dialogue utterances and ontology in the validation, test, and few-shot training data sets to English. We provide the translators with an annotation tool (Figure \ref{fig:UI}) to navigate through data examples, perform translation, and highlight entity spans in the translated sentence. The tool helps verify the consistency of slot value translations between user/agent utterances and their annotations after translation. 

For each utterance in a dialogue, our tool automatically identifies the values in dialogue states and user/agent actions. Slots are {\em canonicalized} before calling the database, meaning that their values must lexically match those in the ontology.
Since slot values appearing in the utterances may differ from the canonicalized version, we ask translators to manually identify and mark the non-canonicalized form of slot values and their word spans in the utterances.

The tool automatically checks the number of highlighted spans to prevent missing entity translations. After checking, the annotation tool outputs the English dialogue texts and a correspondence (i.e. alignment) between source and target language slot values.

\subsubsection{From English to Other Languages}
\label{sec:auto-translate}

\paragraph{Automatic Translation.}
For validation, test, and few-shot data, we use commercial translation models since (1) translation is done only once, (2) data size is smaller so it is affordable, and (3) higher data quality reduces post-editing effort.

\paragraph{Manual Post Editing.}
We hire bilingual speakers of English and the target language to post-edit the translations for fluency and correctness. We instruct them to update the alignment if they modify the translated entities. We provide several tools that automatically check their work and help them during the process. We describe the details in Section~\ref{sub:verify}.

\subsection{Zero-Shot Training Data Translation \& Alignment}
To create the zero-shot training datasets for the target languages (including English), we use open-source machine translation models to translate the Chinese data to the target language. We pick open-source models since (1) their results are reproducible, (2) open-source models provide access to model weights necessary for hybrid alignment (described below), (3) they allow tuning text generation hyperparameters such as temperature~\cite{ficler2017controlling} or beam size~\cite{freitag2017beam} and (4) they cost less, thus allowing effective scaling to more languages.

\paragraph{Hybrid Alignment for NMT.}
Previous work~\cite{moradshahi2021contextual,li-etal-2021-mtop} proposed using alignment for tracking the position of entities during translation to ensure they can be replaced with the desired translation both in the utterance and the belief state.
For this purpose, the encoder-decoder cross-attention weights of the neural machine translation model were used in a method called {\em neural alignment}. Although neural alignment often works well, it can produce incorrect spans as it is a probabilistic approach and has particularly low recall on long multi-token entities. 

Ideally, if there exists a dictionary that provides a mapping between each source entity and all possible translations in the target language, we can directly scan the translated sentence to see if there is a match. We call such an approach {\em dictionary alignment}. Unfortunately, there is no such dictionary. 
We propose to build such a dictionary for each sentence on-the-fly. To do so, we first extract the entities from the sentence, then translate each individually and use nucleus sampling~\cite{holtzman2019curious} with different temperature values to generate $K$ translation candidates. This way, we build a mapping between each entity and possible translations which serves as the dictionary for dictionary alignment. Finally, we combine the two methods in a {\em hybrid} approach: We try to use dictionary alignment first, and if there is no matching translation in the output, we fall back to neural alignment.

\subsection{Creating English-Hindi Code-Mixed Zero-Shot Training Data}
For generating English-Hindi code-mixed train set, we implemented a pipeline combining GCM~\citep{rizvi-etal-2021-gcm}, and alignment based word substitution. An overview of the pipeline is shown in Fig.~\ref{fig:enhi-trainset}. GCM automatically generates code-mixed text given parallel data in two languages, based on two linguistic theories of code-mixing, the Equivalence Constraint theory~\citep{poplack-article} and the Matrix Language theory~\citep{scotton1993duelling}.

We take the Chinese training set as source and translate user and agent utterances to English (en) and Hindi (hi). The translated sentences are fed as input to GCM, which produces code-mix utterances. For sentences where GCM fails to generate any candidate, we rely on word-alignment-based word substitution to generate a code-mixed utterance. Alignments are generated using cosine similarities between sub-word representations from mBERT in a parallel sentence pair~\citep{dou-neubig-2021-word}.

\subsection{Translation of Annotations}

The next step is to translate the slot values in the belief state, user and agent acts, and database search results in the source language to the target language. Since the translations of the same slot value may vary according to the context (e.g., ``\chin{是}'' corresponds to is, does, has or other words indicating affirmative), we create a one-to-many mapping between source language slot values and corresponding translations based on the slot value alignments obtained above. 
We ask human translators to select the most appropriate expression from all candidate translations as the canonicalized translation. We follow two basic principles in this process:

{\bf Part-of-Speech (POS) Consistency.} The translator should pick, for each slot, values with the same POS tags where possible. For example, for the ``production country/region'' slot in the TV series domain, we will use the unified noun form (i.e., ``America''/``India'') instead of mixing the noun and adjective form (i.e., ``American''/``India'').

{\bf Value Consistency.} The translator should use the same translation across domains and slots. For example, the Chinese word ``\chin{中等}'' when used as a ``price-range'' can be translated into ``moderate'' or ``medium''. We consistently map ``\chin{中等}'' to ``moderate'' for all ``price-range'' slots across all domains.

\subsubsection{Creating Ontology and Databases} 
We found that ontology construction should be done in tandem with dataset translation. In prior work, using a predefined ontology limited fluency and diversity of the translations~\cite{zuo2021allwoz}, and replacing entities in sentences after translation without careful attention to parts of speech or context resulted in grammatically incorrect sentences~\cite{moradshahi-etal-2020-localizing,ding-etal-2022-globalwoz}. Each value in the source database is automatically mapped to its canonicalized translation. Note that since not all slot values are seen in the training dataset, translators are asked to provide canonicalized translations for those values.

The original RiSAWOZ dataset only provides final search results from databases instead of intermediate API calls. We hence also restore the API calls through the dialogue state, database, and search results for complete database interactions. This improves the extensibility of the dataset and helps to generalize RiSAWOZ to other languages and domains in the future.

\subsubsection{Annotation Checker}
\label{sub:verify}
Manual errors are inevitable, especially for translators who are unfamiliar with the process. We have developed an annotation checker to automatically flag and correct errors where possible:

{\bf Entity Checking.}
Our annotation checker ensures that changes made in the English translation of entities are propagated to the downstream translation for other target languages. It compares the revised annotations with current annotations and deleted incorrect or redundant slots. Additionally, it locates missing entities or entities that need reannotation to help annotators quickly synchronize the latest changes.

{\bf API Checking.}
Some datasets such as RiSAWOZ, include the ground truth database search results. For these datasets, we can check the consistency of the API by comparing the results of the API call with the provided ground truth. Our checker resolves observed discrepancies by automatically deleting redundant slots and values in constraints and adding the differences to the slot value mappings. It also shows the precise locations of changes for annotators to review.

\section{Experiment}
\label{experiment}

The goal of our experiments is to create an agent in a {\em target} language, given full training data in the source (Chinese) language, and a varying amount of training data in the target language. We also assume we have access to a machine translation model from Chinese to the target language. We perform our experiments on different target languages in X-RiSAWOZ. Table~\ref{tab:data_stats} shows statistics of different data splits used in the experiments, which is the same across all {\em target} languages.

\subsection{Setting}
\paragraph{Full-Shot (mono-lingual).}
This setting is only possible for Chinese since we do not have full training data for  target languages. In the full-shot experiments, all of the original Chinese training data is used for training. Note that this setting is not a cross-lingual experiment per se, but a point of comparison for other settings.

\paragraph{Zero-Shot (cross-lingual).}
In our zero-shot experiments, no manually created target language data is available for training. Instead, we automatically create training data by machine translation of the source language as described in Section~\ref{sec:auto-translate}. Additionally, we perform two ablations on our automatic training data translation approach: (1) Only using neural alignment ($-$ Dictionary Align) (2) No alignment of any type ($-$ Neural Align).

\paragraph{Few-Shot (cross-lingual).}
In the few-shot setting, we start from a zero-shot model (with its various ablations) and further fine-tune it on the few-shot dataset in the target language. So the model is trained on both machine translated data and few-shot manually created dataset. In this setting, we also perform an ablation where we only train on the few-shot training data and no machine translated data ({\em Few-shot Only}).

\begin{table*}
\small
\centering
\newcolumntype{C}{>{\centering\arraybackslash}X}
\resizebox{1.85\columnwidth}{!}{
\begin{tabularx}{\linewidth}{l|p{2.5cm}|C|C|C}

\specialrule{.1em}{.0em}{.0em} 

Language & Setting & DST Acc. $\uparrow$ & DA Acc. $\uparrow$  & BLEU $\uparrow$  \\

\specialrule{.1em}{.0em}{.0em} 
\multicolumn{5}{c}{\em Full-Shot} \\
\specialrule{.1em}{.0em}{.0em} 
\multirow{1}{*}{Chinese} & Ours  & \bf 96.43 & \bf 91.74 & \bf 51.99 \\

\specialrule{.1em}{.0em}{.0em} 
\multicolumn{5}{c}{\em Zero-Shot} \\
\specialrule{.1em}{.0em}{.0em} 
\multirow{3}{*}{English} & Ours & \bf 84.23 & \bf 67.27 &  \bf 27.14   \\
& $-$ Dictionary Align  & 83.42 & 66.51 & 22.67 \\
& $\quad - $ Neural Align & 82.33 & 67.79 & 13.24 \\
\hline
\multirow{3}{*}{French} & Ours  & \bf 70.75 & \bf 59.27 & \bf 29.88 \\
& $-$ Dictionary Align  & 68.22  &  56.32  & 25.43 \\
& $\quad - $ Neural Align    &  64.53 & 53.33   &  18.12  \\
\hline
\multirow{3}{*}{Hindi} & Ours  & \bf 52.09 & \bf 56.06 & \bf 27.42 \\
& $-$ Dictionary Align   & 50.12  & 54.34   & 23.43  \\
& $\quad - $ Neural Align   &  48.11 & 53.21   &  18.32  \\
\hline
\multirow{3}{*}{Korean} & Ours   & \bf 34.55 & 49.56 & \bf 10.17 \\
& $-$ Dictionary Align   & 31.47 & \bf 50.17 & 9.87 \\
& $\quad - $ Neural Align  & 29.87 & 49.51 & 4.59 \\
\hline
\multirow{1}{*}{English-Hindi} & Ours   & \bf 49.95 & \bf 42.78 & \bf 11.31 \\

\specialrule{.1em}{.0em}{.0em} 
\multicolumn{5}{c}{\em Few-Shot} \\
\specialrule{.1em}{.0em}{.0em} 
\multirow{1}{*}{Chinese} & Few-shot Only    & \bf 82.75 & \bf 77.33 & \bf 38.87  \\
\hline
\multirow{4}{*}{English} & Ours    &  \bf 84.62 &  69.44 &  \bf 46.37   \\
& $-$ Dictionary Align   & 83.37 & 69.74 & 46.16 \\
& $\quad - $ Neural Align      & 82.01 &  \bf  70.45 & 45.43 \\
& Few-shot Only   & 74.52 & 58.97 & 45.53 \\
\hline
\multirow{4}{*}{French} & Ours   &  \bf 73.12  & \bf 61.11   &  \bf 42.21  \\
& $-$ Dictionary Align     & 71.12 & 60.21  & 40.12  \\
& $\quad - $ Neural Align     & 69.68 & 57.12  & 38.14  \\
& Few-shot Only   & 67.55 & 50.96 & 44.77    \\
\hline
\multirow{4}{*}{Hindi} & Ours     & 75.16 & \bf 59.02 & \bf 38.38  \\
& $-$ Dictionary Align     & \bf 75.32  & 57.66   & 37.54   \\
& $\quad - $ Neural Align      & 73.21  & 54.32   & 34.32   \\
& Few-shot Only    & 55.77 & 49.88 & 38.18  \\
\hline
\multirow{4}{*}{Korean} & Ours       & \bf 71.17 & \bf 53.52 & \bf 34.93   \\
& $-$ Dictionary Align     & 69.57 & 52.37 & 34.75 \\
& $\quad - $ Neural Align      & 69.91 & 52.00 & 33.80  \\
& Few-shot Only     & 60.65 & 41.47 & 32.76   \\
\hline
\multirow{2}{*}{English-Hindi} & Ours      & \bf 60.67 & \bf 37.97 & 26.77   \\
& Few-shot Only     & 56.53 & 36.50 & \bf 28.54  \\

\specialrule{.1em}{.0em}{.0em}
\end{tabularx}
}
\vspace{-0.8em}
\caption{Results on the validation set of \xwoz, obtained by feeding the gold input for each subtask in each turn. The best result in each section is in bold. $\uparrow$  indicates higher number shows better performance.}
\label{tab:gold-accuracies}
\vspace{-1.7em}
\end{table*}

\subsection{Models}
In all our experiments, we use the m2m100 ~\cite{fan2020beyond} model for Korean and mBART~\cite{liu2020multilingual} for all other languages. We found mBART to be especially effective in zero-shot settings as the language of its outputs can be controlled by providing a language-specific token at the beginning of decoding. Additionally, its denoising pre-training objective improves its robustness to the remaining translation noise. In each setting, all four dialogue subtasks are done with a single model, where we specify the task by prepending a special token to the input.

Since the dataset for target languages is introduced in this paper, there is only prior work on the Chinese dataset. In Section~\ref{sec:FullEvaluation}, we compare our results to the best previously reported result on RiSAWOZ from \citet{moradshahi2021contextual} that achieved SOTA on the DST subtask using an mBART model, and from \citet{risawoz} for other subtasks which use DAMD~\cite{zhang2020task}, a Seq2Seq RNN end-to-end dialogue model. We use seven widely-used automatic metrics to compare different models. Please see Section~\ref{sec:metrics} for details of each metric.

\section{Results and Discussion}

\begin{table*}[!htb]
\fontsize{8}{10}\selectfont
\small
\centering
\newcolumntype{C}{>{\centering\arraybackslash}X}
\resizebox{1.85\columnwidth}{!}{
\begin{tabularx}{\linewidth}{l|p{2.5cm}|C|C|C|C|C|C|C}

\specialrule{.1em}{.0em}{.0em} 

Language & Setting & JGA $\uparrow$ & TSR $\uparrow$  & DSR $\uparrow$  & API $\uparrow$ &  DAA $\uparrow$  & BLEU $\uparrow$  & SER $\downarrow$ \\

\specialrule{.1em}{.0em}{.0em} 
\multicolumn{9}{c}{\em Full-Shot} \\
\specialrule{.1em}{.0em}{.0em} 
\multirow{2}{*}{Chinese} & Ours  & \bf 78.23 & \bf 53.67 & \bf 45.67 & \bf 72.70 & \bf 73.68 & \bf 34.72 & \bf 26.41 \\
& SOTA & 76.90 & 48.63 & 40.32 & -- & -- & 27.90 & 30.32 \\

\specialrule{.1em}{.0em}{.0em} 
\multicolumn{9}{c}{\em Zero-Shot} \\
\specialrule{.1em}{.0em}{.0em} 
\multirow{3}{*}{English} & Ours  & \bf 43.64 & \bf 22.46 & \bf 16.00 & \bf 44.95 & \bf 40.81 & \bf 14.12 & \bf 47.08 \\
& $-$ Dictionary Align   & 38.70 & 19.22 & 13.50 & 39.84 & 37.35 & 11.34 & 49.64 \\
& $\quad-$ Neural Align    & 38.96 & 9.50 & 5.67 & 40.95 & 41.96 & 8.21 & 59.90\\
\hline
\multirow{3}{*}{French} & Ours  & \bf 24.04 & \bf 12.58 & \bf 7.17 & \bf 34.20 & \bf 38.32 & \bf 10.88 & \bf 58.45  \\
& $-$ Dictionary Align   & 20.32 & 5.43 & 4.18 & 28.51 & 35.78 & 9.72 & 60.25  \\
& $\quad-$ Neural Align    & 19.43 & 3.23 & 2.11 & 24.64 & 28.36 & 6.81 & 68.89 \\
\hline
\multirow{3}{*}{Hindi} & Ours     & \bf 20.32  & \bf 10.11  & \bf 4.32  & \bf 32.32 & \bf 34.23  & \bf 9.13  & \bf 60.43  \\
& $-$ Dictionary Align    &  18.31 & 5.15  & 3.98  & 30.12 & 32.31  & 8.11  & 65.43  \\
& $\quad-$ Neural Align     & 17.32  & 3.12  & 3.10  & 28.51 & 28.13  & 7.00  & 67.25  \\
\hline
\multirow{3}{*}{Korean} & Ours      &  \bf  21.41 & \bf 10.75 & \bf 5.00 & \bf 32.08 & \bf 36.57 & 7.27 & \bf 64.33 \\
& $-$ Dictionary Align     & 19.53 & 9.46 & 4.83 & 27.75 & 36.33 & \bf  7.55 & 35.84   \\
& $\quad-$ Neural Align    & 17.77 & 8.77 & 3.67 & 27.19 & 25.45 & 7.12 & 38.98   \\
\hline
\multirow{1}{*}{English-Hindi} & Ours   & \bf 9.22 & \bf 4.81 & \bf 2.03 & \bf 10.43 & \bf 26.47 & \bf 5.41 & \bf 63.26 \\

\specialrule{.1em}{.0em}{.0em} 
\multicolumn{9}{c}{\em Few-Shot} \\
\specialrule{.1em}{.0em}{.0em} 
\multirow{1}{*}{Chinese} & Few-shot Only  & \bf 37.69 & \bf 28.04 & \bf 21.00 & \bf 40.73 & \bf 42.30 & \bf 13.89 & \bf 45.44 \\
\hline
\multirow{4}{*}{English} & Ours   & \bf 48.91 & \bf 23.13 & \bf 17.17 & \bf 50.06 & \bf 42.45 & \bf 26.33 & \bf 44.93 \\
& $-$ Dictionary Align  & 48.40 & 22.79 & 16.67 & 50.03 & 42.26 & 25.29 & 45.01 \\
& $\quad-$ Neural Align    & 46.31 & 22.68 & 16.50 & 47.61 & 42.54 & 25.78 & 44.78 \\
& Few-shot Only   & 29.87 & 16.09 & 10.50 & 32.30 & 30.45 & 20.00 & 52.79 \\
\hline
\multirow{4}{*}{French} & Ours    & \bf 30.85 & \bf 17.17 & \bf 11.83 & \bf 39.97 & \bf 45.03 & \bf 20.92 & \bf 46.26 \\
& $-$ Dictionary Align   & 28.51 & 16.11 & 9.54 & 38.11 & 43.41 & 19.91 & 48.35   \\
& $\quad-$ Neural Align    & 26.45 & 15.54 & 9.13 & 35.74 & 42.15 & 16.99 & 49.26   \\
& Few-shot Only   & 19.43 & 3.23 & 2.11 & 24.64 & 28.36 & 6.81 & 68.89 \\
\hline
\multirow{4}{*}{Hindi} & Ours   & \bf 25.62   & \bf 15.67  & \bf 11.31  & \bf 37.54  & \bf 41.32  & \bf 18.51  & \bf 44.26  \\
& $-$ Dictionary Align    & 23.12  & 15.11  & 10.32  & 35.14 & 39.51  & 16.34  & 46.76  \\
& $\quad-$ Neural Align     & 21.12  & 13.22  & 8.61  & 34.11 & 34.12  & 15.33  & 48.97  \\
& Few-shot Only   & 18.48 & 8.16 & 4.50 & 19.09 & 23.41 & 13.15 & 62.24  \\
\hline
\multirow{4}{*}{Korean} & Ours   & \bf 26.24  & \bf 14.32  & \bf 10.60  & \bf 35.42 & \bf 38.42  & \bf 20.32  &  \bf 43.21 \\
& $-$ Dictionary Align    & 24.13  & 12.53  & 8.45  & 23.42 & 33.34  & 19.32  & 47.32  \\
& $\quad-$ Neural Align      & 23.54  & 10.23  & 7.54  & 22.31 & 30.42  & 18.34  & 50.33  \\
& Few-shot Only   & 20.66 & 9.16 & 5.17 & 19.39 & 23.56 & 17.81 & 54.57 \\
\hline
\multirow{2}{*}{English-Hindi} & Ours   & \bf 21.80 & \bf 4.13 & 1.83 & \bf 22.64 & \bf 21.69 & \bf 5.29 & \bf 66.31 \\
& Few-shot Only    & 16.07 & 3.69 & \bf 2.33 & 15.65 & 16.97 & 3.93 & 69.61 \\

\specialrule{.1em}{.0em}{.0em}               
\end{tabularx}
}
\vspace{-0.5em}
\caption{End-to-end results and ablations on the test set of \xwoz. The best result in each section is in bold. $\downarrow$  indicates lower number shows better performance and vice versa.}
\label{tab:e2e-accuracies}
\vspace{-1.5em}
\end{table*}

We first evaluate the models for each turn, assuming that all previous subtasks and steps are correct. We then evaluate the end-to-end accuracy for the whole conversation. 

\subsection{Turn by Turn Evaluation}
To understand how each component performs independently, our first experiment uses the gold data of all the previous turns and subtasks as input in our evaluation (Table~\ref{tab:gold-accuracies}). In this scenario, errors do not propagate from one subtask to the next in each turn. \textit{Ours} refers to our main approach, which combines all techniques. Each ablation incrementally takes away one of the techniques. 

In the zero-shot setting, results vary across added languages, where the agent achieves between 34.6-84.2\% on DST, 42.8-67.3\% on DA, and 10.2-29.9\% on BLEU score. Fine-tuning on the few-shot data improves all metrics for all languages, with the agent achieving between 60.7-84.6\% on DST, 38.0-70.5\% on DA, and 28.5-46.4\% on BLEU score. The improvement in DST is particularly prominent for Hindi, Korean, and English-Hindi, where the quality of machine translation may not be as good. Nonetheless, adding automatically translated data to training greatly improves the accuracy for these languages over the ``few-shot only'' result.

\subsection{Error Analysis}
\label{sec:errors}
To better understand the inference limitations of our trained agents, we manually inspected the model predictions by randomly selecting 100 validation turns for each domain where the prediction was incorrect. The following are the most common error patterns we observed across all languages:

\noindent{\bf Implicit Entities}: 
In \xwoz dialogues, some entities are not mentioned explicitly in the user's utterance and need to be \emph{inferred}. These entities include the corresponding price range for a luxury diner, a speaker's desired attraction for a date with their partner, and hotel rating. These errors are partly due to the limited common-sense capability of the pre-trained language model used~\cite{zhou2020evaluating} and partly due to the training data encouraging the model to copy entities verbatim from the input instead of making logical reasoning. This category accounts for 27\% of errors observed.

\noindent{\bf Multiple Correct Dialogue Acts}: 
In \xwoz, the agent often provides an answer as soon as it receives the API call results. However, in some cases, the agent asks follow-up questions (e.g., ``how many seats do you want for the car?'') to narrow down the search results. Since the dataset is constructed via human interactions and not simulation, there are no well-defined policies governing the agent's behavior. Thus, there are examples where multiple dialogue acts can be correct given the input and API constraints. Since during evaluation we can only check the model output against the one response provided as gold, another perfectly fine response can be deemed as incorrect. We discovered that 38\% of errors are of this nature.

\noindent{\bf Incorrect Entities}: In DST and DA subtasks, the accuracy is highly dependent on identifying the correct entities in the input. However, there are cases where the model (1) predicts a wrong entity, (2) predicts part of the entity, (3) predicts the entity along with prepositions, articles, etc. (4) omits the entity, or (5) fully hallucinates an entity. We found (1) and (2) to be the most common patterns. (3) can be addressed by a simple pre-processing or text lemmatization. (4) happens with complex sentences with many entities where the model often mispredicts the slot names as well as slot values. (5) is usually caused by data mis-annotations or errors in data processing, where a slot is missing from the input and the model generates the most probable value for it. The remaining 35\% of errors fall under this category.

For each language, we also performed a similar analysis to understand if there are language-specific attributes that affect the accuracy and quality of the translated datasets. The result of these analyses is included in the appendix (\ref{sec:appendix-error-analysis-french}-\ref{sec:en-hi}).

\subsection{Full Conversation Evaluation}
\label{sec:FullEvaluation}
The main results of our experiments are reported in Table~\ref{tab:e2e-accuracies}. Following~\citet{lin2021bitod}, the evaluation for these experiments is performed end-to-end meaning for each turn, the model output from the previous subtask is used as input for the next subtask. This reflects a real-world scenario where an agent is conversing with the user interactively. 

Overall, in the full-shot setting, when training on the Chinese dataset, we improve the state of the art in Joint Goal Accuracy (JGA) by 1.33\%, Task Success Rate (TSR) by 5.04\%, Dialogue Success Rate (DSR) by 5.35\%, and BLEU by 6.82\%.
The improvements are due to the improved and succinct dialogue representation we have created (Section~\ref{sec:interface}), and contextual representations of transformer models.  

In the zero-shot setting, results vary across languages, where the English, French, Hindi, Korean, and English-Hindi agents achieve 35\%, 16\%, 9\%, 11\%, and 4\% of the DSR score of the full-shot Chinese agent, respectively. In the few-shot setting, the ratio improves to 38\%, 26\%, 25\%, 23\%, and 5\%. The smallest and biggest improvements are on the English and Hindi dataset respectively. This suggests that the impact of few-shot data is greater when the quality of the pretraining data is lower, which is related to the quality of the translation model between Chinese and the target language.

 The Response Generation subtask receives the largest improvement in performance when provided with human supervision in the few-shot data, with a BLEU score improvement of over 10\%. This suggests that while translation with alignment is effective for understanding user input, it is not as effective for generating output text. This is partly due to the agent model used, mBART, which is trained with a denoising objective and is thus able to handle noisy input text better.

\section{Conclusion}
\label{conclusion}

This paper presents a solution for balancing the trade-offs between standard machine translation and human post-editing. By standardizing and establishing best practices for ``translation with manual post-editing'', and releasing associated toolkits, post-editing can be made faster, more efficient, and cost-effective. We use our methodology to create \xwoz, a new end-to-end, high-quality, and large multi-domain multilingual dialogue dataset, covering 5 diverse languages and 1 code-mixed language. We also provide strong baselines for zero/few-shot creation of dialogue agents via cross-lingual transfer. In the few-shot setting, our agents achieve between 60.7-84.6\% on DST, 38.0-70.5\% on DA, and 28.5-46.4\% on RG subtasks across different languages. Overall, our work paves the way for more efficient and cost-effective development of multilingual task-oriented dialogue systems.

\section{Limitations}

We would have liked to evaluate the generalization of our cross-lingual approach on more languages. For instance, we partially rely on machine translation models for Chinese-to-English translation. Available translation models for other language pairs, especially from/to low-resource languages have much lower quality, and it would be desirable to measure the effect of that in our experiments.

The ontology used for new languages is derived by translating the Chinese ontology. As a result, the entities are not localized. Creating local ontology requires manual effort as one would need to identify websites or databases for scraping or collecting the entities. Once the local entities are collected, we can automatically replace translated entities with local ones to localize the dataset.

Another limitation is the lack of human evaluation for agent responses. BLEU score does not correlate well with human judgment~\cite{sulem2018bleu}, and SER only accounts for the factuality of the response but not grammar or fluency. In future work, we wish to address this by conducting human evaluations in addition to automatic metrics.

\section{Ethical Considerations}
\label{sec:ethics}

We do not foresee any harmful or malicious misuse of the technology developed in this work. The data used to train models is about seeking information about domains like restaurants, hotels and tourist attractions, does not contain any offensive content, and is not unfair or biased against any demographic. This work does focus on widely-spoken languages, but we think the cross-lingual approach we proposed can improve future dialogue language technologies for a wider range of languages.

We fine-tune multiple medium-sized (several hundred million parameters) neural networks for our experiments. We took several measures to avoid wasted computation, like performing one run instead of averaging multiple runs (since the numerical difference between different models is large enough to draw meaningful conclusions), and improving batching and representation that improved training speed, and reduced needed GPU time.
Please refer to Appendix~\ref{sec:implementation} for more details about the amount of computation used in this paper.

\section*{Acknowledgements}
We would like to thank Ruchi Jain for helping us validate the automatically translated Hindi dialogues.
This work is supported in part by the National Science Foundation
under Grant No.~1900638, the Alfred P. Sloan Foundation under Grant No.~G-2020-13938, the Verdant Foundation, Microsoft, KDDI, JPMorgan Chase, and the Stanford Human-Centered Artificial Intelligence (HAI) Institute.
This work is also co-funded by the Natural Science Foundation of Xinjiang Uygur Autonomous Region (No. 2022D01D43), Zhejiang Lab (No. 2022KH0AB01). This project has also received funding from the European Union’s Horizon 2020 Research and Innovation Programme under Grant Agreement No. 101021797 (Starlight), and the European Union’s Horizon Europe research and innovation programme under grant agreement N° 101070192 (CORTEX²). 
This work is also supported in part by Institute of Information \& communications Technology Planning \& Evaluation (IITP) grant funded by the Korea government(MSIT) (No.2020-0-01373 and IITP-2022-2021-0-01817).

\bibliography{anthology,custom}

\appendix
\section{Appendix}
\label{sec:appendix}

\subsection{Implementation details}
\label{sec:implementation}
Our code is implemented in PyTorch~\cite{paszke2019pytorch} using GenieNLP~\cite{geniepldi19}
~\footnote{\href{https://github.com/stanford-oval/genienlp}{https://github.com/stanford-oval/genienlp}}
library for training and evaluation. We use our newly written library (described in Section~\ref{sec:interface}) for data preprocessing and evaluation which will be released upon publication. We use pre-trained models available through HuggingFace's Transformers library~\cite{Wolf2019HuggingFacesTS}. We use \textit{m2m100-418M} model for Korean and \textit{mbart-large-50} for other languages as the neural model for our agent. Both models use a standard Seq2Seq architecture with a bidirectional encoder and left-to-right autoregressive decoder. mBART uses sentence-piece~\cite{kudo2018sentencepiece} for tokenization, and is pre-trained on text reconstruction task in 50 languages.

In each setting, all four dialogue subtasks are done with a single model, where we specify the task by prepending a special token to the input. We found mBART to be especially effective in zero-shot settings as the language of its outputs can be controlled by providing a language-specific token at the beginning of decoding. Additionally, its denoising pre-training objective improves its robustness to the remaining translation noise.

For translation, we use the publicly available \textit{mbart-large-50-many-to-many-mmt} (\textasciitilde611M parameters) and \textit{m2m100-418M} (\textasciitilde1.2B parameters) models which can directly translate text from any of the 50 supported languages.
 
 We use greedy decoding and train our models using teacher-forcing and token-level cross-entropy loss. We use Adam~\cite{kingma2014adam} as our optimizer with a start learning rate of $2 \times 10^{-5}$ and linear scheduling. These hyperparameters were chosen based on a limited hyperparameter search on the validation set. For the numbers reported in the paper, due to cost, we performed only a single run for each experiment.
 
Our models were trained on virtual machines with a single NVIDIA V100 (16GB memory) GPU on the Azure platform. For a fair comparison, all models were trained for the same number of iterations of 200K in the full-shot setting. In the few-shot setting, we fine-tuned the model for 10K steps on the few-shot data.
Sentences are batched based on their input and approximate output token count for better GPU utilization. We set the total number of tokens per batch to 720. Training and evaluating each model takes about 15 GPU hours on average.
 
At inference time, we use the predicted belief state as input to subsequent turns instead of ground truth. However, to avoid the conversation from diverging from its original direction, similar to ~\citet{lin2021bitod}, we use ground-truth agent acts as input for the next turn. We made sure the settings are equivalent for a fair comparison. Additionally, we noted that in many examples the prediction is similar to the gold truth except for small differences such as in case~(e.g., ``district'' vs ``District''), or extra punctuation in the predicted output. To address this, during evaluation, we apply entity normalization by using canonical mapping and string pattern matching to map entities to their canonicalized form.

\subsection{Evaluation Metrics}
\label{sec:metrics}
Following~\citet{moradshahi-etal-2023-zero}, we use the following metrics to compare different models. Scores are averaged over all turns unless specified otherwise. 

\begin{itemize}[leftmargin=*,topsep=0pt,noitemsep]
    \item
    \textbf{Joint Goal Accuracy (JGA)}~\cite{multiwoz1}: The standard metric for evaluating DST. JGA for a dialogue turn is 1 if all slot-relation-value triplets in the generated belief state match the gold annotation, and is 0 otherwise.
    
    \item
    \textbf{Task Success Rate (TSR)}~\cite{lin2021bitod}: A task, defined as a pair of domain and intent, is completed successfully if the agent correctly provides all the user-requested information and satisfies the user's initial goal for that task. TSR is reported as an average over all tasks. 
    
    \item
    \textbf{Dialogue Success Rate (DSR)}~\cite{lin2021bitod}: DSR is 1 for a dialogue if all user requests are completed successfully, and 0 otherwise. DSR is reported as an average over all dialogues. We use this as the main metric to compare models, since the agent needs to complete all dialogue subtasks correctly to obtain a full score on DSR.
    
    \item
    \textbf{API}: For a dialogue turn, is 1 if the model correctly predicts to make an API call, and all the constraints provided for the call match the gold. It is 0 otherwise.
    
    \item
    \textbf{Dialogue Act Accuracy (DAA)}: For a dialogue turn, is 1 if the model correctly predicts all the dialogue acts including entities, and is 0 otherwise. 
    
    \item
    \textbf{BLEU}~\cite{papineni2002bleu}: Measures the natural language response fluency based on n-gram matching with the human-written gold response. BLUE is calculated at the corpus level.
    
    \item
    \textbf{Slot Error Rate (SER)}~\cite{wen2015semantically}: It complements BLEU as it measures the factual correctness of natural language responses. For each turn, it is 1 if the response contains all entities present in the gold response, and is 0 otherwise.
\end{itemize}

\subsection{Human Post-Editing}

\begin{figure*}
    \centering
    \includegraphics[width=0.7\linewidth]{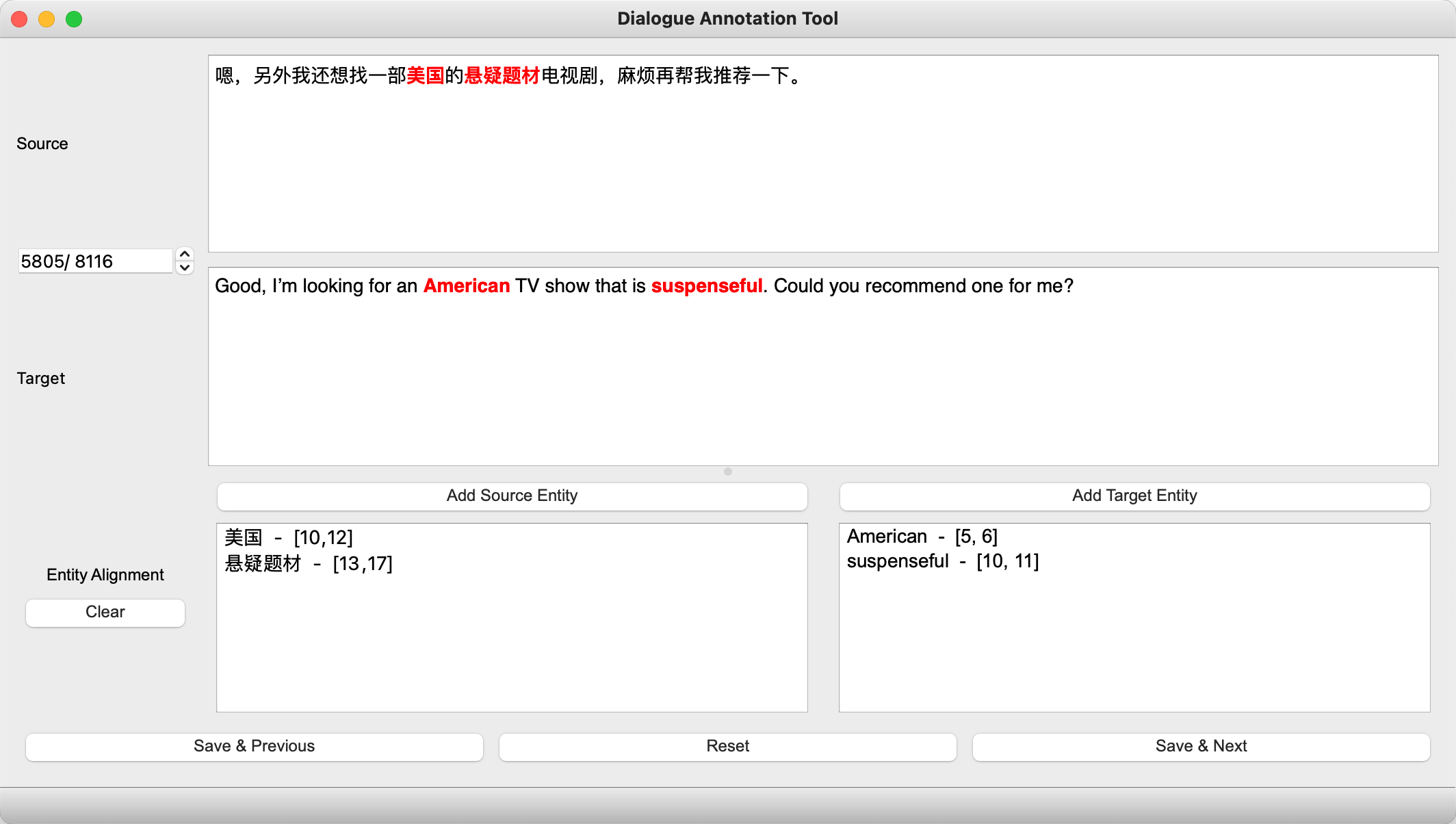}
    \caption{A screenshot of the annotation tool used by translators to translate a sentence from Chinese to English and mark the entity spans to create the slot value alignment. The entity spans show the position of words for English and characters for Chinese.}
    \label{fig:UI}
\end{figure*}

Bilingual speakers of the source and output language were recruited as human translators and post-editors by each team. A user interface (see Fig.~\ref{fig:UI}) was provided for them to perform translation and alignment tasks. The translators were instructed to ensure that the resulting translations were both accurate and fluent. Compensation for their work was provided at the standard rate in their respective countries.

\subsection{Error Analysis: French}
\label{sec:appendix-error-analysis-french}

    For the French language, we focused on the Response Generation subtask. We selected 300 of the prediction examples marked as false as not matching exactly the reference. In this set 42.8\% of the predictions are completely wrong. Polite forms are particularly problematic. As an example, we can cite the case of the expression \textit{``Tout le plaisir est pour moi, au revoir. }(\textit{It's my pleasure, goodbye.})'' which has four different wrong predictions \textit{``Pas de courtoisie, au revoir.} (\textit{No courtesy, goodbye.})'', \textit{``Pas de gentillesse, au revoir.} (\textit{No kindness, goodbye.})'', \textit{``Je suis heureux de vous servir, au revoir.} (\textit{I am pleased to serve you, goodbye.})'' and \textit{``Pas de bonheur, adieu! }(\textit{No happiness, goodbye})''. 
The root cause most likely stems from the literal translation of Chinese idioms used in polite expressions (\chin{不要客气}) to French in the zero-shot training data. However, most of the polite expressions should be easy enough to correct.

We noted that 7.4\% of the predictions are just slightly off semantically. For example (\textit{``Je recommande l'université de Xi'an Jiaotong-Liverpool.} (\textit{I recommend Xi'an Jiaotong-Liverpool University.})''  vs. \textit{``l'université de Xi'an Jiaotong-Liverpool de Liverpool est très bien.} (\textit{Xi'an Jiaotong-Liverpool University from Liverpool is very good.})'' with the wrong insertion of \textit{``de Liverpool} (\textit{from Liverpool})''. 

On the other hand, 15.4\% of the predictions are semantically correct, but with minor errors (syntactic errors, repetitions, etc.). For instance, the meaning of the sentence \textit{``Il y aura une brise sans direction continue samedi prochain.} (\textit{There will be a continuous directionless breeze next Saturday.})'' is the same than the one of the sequence of words \textit{``Le vent une brise sans direction continue vent doit être doux.} (\textit{The wind a breeze without continuous wind direction must be gentle.})'' but this sequence of words is syntactically wrong. The date reference is also missing but it was not mandatory for the correctness of the dialogue. Finally, 34.4\% of the supposedly wrong generated responses are in fact correct but just expressed differently, like in \textit{``Elle a une note de 4,3.} (\textit{It has a rating of 4.3.}'') vs. \textit{``La note de ce lieu est de 4,3.} (\textit{The rating for this location is 4.3.})''. We think that this kind of difference could be handled by a computation of sentence embedding distance.

As we focused on the Response Generation component, we did not carry out a large-scale qualitative analysis of the Slot-Relation subtask but a quick look at the data seems to indicate that some given slots are often missing from the generated part, like \textit{"the\_most\_suitable\_people"}. For some other slots like ``\textit{metro\_station}'', some values seem to be missing from normalization data like ``\textit{peut}'' which should be equivalent to ``\textit{pouvoir}'' and ``\textit{true}''. This latter error will be quite simple to correct.

\subsection{Error Analysis: Hindi}
We sample 10\% of the errors from each domain from the Hindi validation dataset and analyze these examples manually. The following are the error patterns we observe:

\textbf{Response Generation}. As discussed in Section \ref{sec:en-hi}, there are multiple ways to generate a sentence while matching the semantic content of the gold truth. While such RGs should ideally be marked as correct, their BLEU scores are low. Such instances amount to over 65\% of all RG errors. In addition to such kind of errors, we observe that approximately 18\% of all the RG error samples are largely accurate but they lack fluency. Here is one such instance where the model is trying to say bye to the user: ``ajib hai, alvida!'', which translates to ``That's strange, bye!''. In this example, the model conveys the right message but not in the most polite way. Such instances become more common when the model has to fill the ``general'' slot, used mainly in greetings. This is possibly because the model finds it  more difficult to generate open-ended text than content-guided text. 

\textbf{Erroneous Slot-Relation Values}. In some cases, the model predicts the right slot-relation values, but they are deemed incorrect because it predicts the synonym of the gold truth. This amounts to 28\% of all the erroneous slot-relation value examples. In addition to such instances, we observe that some slot-relation values are marked as incorrect because of minor differences between the gold truth and the model prediction. These include extra spaces,  punctuations, stop words, and the usage of synonyms. Such kind of errors amount to 17.8\% of the sampled erroneous slot relation values. Lastly, our analysis reveals that there seems to be an increased amount of confusion between the following pairs of slots: ``inform'', ``request'' and ``date'', ``time''.

\subsection{Error Analysis: Korean}
The Korean language poses some unique challenges.
In Korean, a word can be made up of multiple characters, and an {\em eo-jeol} is formed by one or more words to convey a coherent meaning.
Spaces are used to delimit an eo-jeol.   For instance, postpositions, or {\em jo-sa} in Korean, are connected to a noun to form an eo-jeol to indicate its grammatical relation to other words in a sentence.

Consider ``\kor{저는 샤먼에 갈거에요}'', a sentence containing 3 eo-jeol and 9 characters that means ``I will go to Xiamen''.  ``\kor{샤먼에}'' is an eo-jeol meaning ``to Xiamen'', where ``\kor{샤먼}'' is ``Xiamen'' and ``\kor{에}'' (which is a jo-sa) means ``to''. Because the two words are connected into a single eo-jeol, the annotation is more prone to mistakes. Furthermore, extracting an entity in an eo-jeol is more difficult. This leads to more ``incorrect entities'' problems in the results for Korean.

Furthermore, Korean possesses distinctive auxiliary verbs/adjectives known as {\em bo-jo yong-eon}. These bo-jo yong-eon can be connected to the main verbs/adjectives either within a single combined eo-jeol or across multiple eo-jeols, leading to similar challenges in entity annotation. 
For example, both ``\kor{친구들과 갈}'' and ``\kor{친구들과 가는}''  means ``to go with friends''. Here, both  ``\kor{-ㄹ}'' (the character at the bottom of ``\kor{갈}'') and ``\kor{는}'' are bo-jo yong-eon meaning ``to go''.
A single English auxiliary verb can map to a wide variety of bo-jo yong-eon depending on the context. 

We modified the annotation tool so that it works at the character level instead of word level. To identify number entities, we use heuristics to extract the jo-sa from eo-jeol composed of a number and a jo-sa. Despite this, our analysis suggests that the eo-jeol and bo-jo yong-eon issues account for approximately 5\% of the errors encountered.

Another issue that we ran into is how negative questions are answered differently in Korean. For example, when asked ``isn't it hot?'', ``yes'' means ``it is hot'' in English, but ``it is not hot'' in Korean.  This discrepancy caused issues during the annotation process. At times, translators mistakenly mapped ``yes'' in English to mean ``no'' in Korean for negative questions, or they transformed them into positive inquiries, which we discovered later on. The former case of mapping ``yes'' to ``no'' resulted in inconsistency in entity mapping, especially when both positive and negative questions are present in the dataset for the domain and slots. To address this, we manually corrected the annotation results to ensure consistent entity mappings, which resolved the majority of the errors.

\subsection{Error Analysis: English-Hindi Code-mixed}
\label{sec:en-hi}
To understand the errors of English-Hindi~(en-hi) code-mix set, we also sampled 10\% of the erroneous examples for each domain from the en-hi validation set. In addition to the error categories noticed for English~(Section~\ref{sec:errors}), we observe the following patterns:

\textbf{Response Generation~(45\%).} Model prediction for Response Generation step is low on BLEU score because there can be multiple ways of code-mixing a sentence. The response could be monolingual, or can be code-mixed to various degrees, or different spans within a sentence could be switched, and such errors account to 19\% of the total errors. For example, the gold truth is ``\textit{yah} 179 minutes \textit{tak chalta hai}'' and the model output is ``movie \textit{ki} duration 179 minutes \textit{hai}''\footnote{In the examples, Hindi tokens~(in italic) are written in romanized format for ease of reading. In the datasets, Hindi tokens are in Devanagri script.}. For around 20\% samples, the generated responses are incoherent, malformed sentences or unnatural code-mixed sentences. We also observed that the generated sentences are low on fluency, while matching the semantic content of the gold truth, accounting for 6\% of total errors. It is our conjecture that the erroneous code-mixed text generation can be ascribed to mBART's restricted ability to generate code-mixed sentences.

\textbf{Erroneous slot-relation-value~(35\%).} In some cases the model predicts additional slot-relation-values, in addition to the correct slot-relation-values~(10\% of the erroneous samples). For example, gold truth is ``$($weather$)$ date equal\_to next Tuesday'' and the predicted output is ``$($weather$)$ city equal\_to Suzhou, date equal\_to next Tuesday''. It is likely that the model is copying additional slot-relation-value tuples that are available in the knowledge part of the input. In 23\% of the analyzed erroneous samples, the model output has the wrong action, domain, slot, relation or slot values. About 1\% of the erroneous samples hallucinated slot values.

\textbf{Language and Script Difference~(20\%).} Across the DST, DA, and RG steps, the gold truth differs from prediction in terms of the script or the language or both. For instance, the slot value could be in Hindi in the Devanagari script, whereas the model prediction is in English or/and in the Roman script. In some cases, although the values match, differences in script/languages can cause the automatic approach to identify them as an error. For example, the gold truth ``$($train$)$ date equal\_to `next Sunday morning' , seat\_type equal\_to `second class ticket' '' differs only slightly from the model output ``(train) date equal\_to `next Sunday morning', seat\_type equal\_to `second class' ''. The measured error rate may not reflect the correct model performance because some of these errors can be reduced by accounting for the semantic match between the generated output and the gold truth.

\subsection{Dialogue Example}

In Table~\ref{tab:examples-dataset}, we show two turns of an example in the original dataset, and its translation to other languages.

\begin{table*}
\begin{tabularx}{\linewidth}{c|c|c|X}
\specialrule{.1em}{.05em}{.05em} 
\multirow{35}{*}{\bf Turn 1} & \multirow{6}{*}{DST} & \multirow{2}{*}{Input (EN)} &  \small DST: <state> null <endofstate> <history> USER: Hi, my friend is coming to Suzhou to visit me, I want to take him to a commercial center in the mid-price range. Do you have anything to recommend? <endofhistory> \\
\cline{3-4}
& & Output (EN) & \small ( attraction ) consumption  " mid " , type  " commercial center " \\
\cline{3-4}
& & Input (ZH) & \small DST: <state> null <endofstate> <history> USER: \chin{你好,我朋友要来苏州找我玩,我想带他找一个消费中等的商业中心逛逛,求推荐。} <endofhistory>\\
\cline{3-4}
& & Output (ZH) & \small ( attraction ) consumption  " \chin{中等} " , type  " \chin{商业中心} " \\
\cline{2-4}
& \multirow{7}{*}{API} & \multirow{2}{*}{Input (EN)} &  \small API: <knowledge> null <endofknowledge> <state> ( attraction ) consumption  " mid " , type  " commercial center " <endofstate> <history> USER: Hi, my friend is coming to Suzhou to visit me, I want to take him to a commercial center in the mid-price range. Do you have anything to recommend? <endofhistory> \\
\cline{3-4}
& & Output (EN) & \small yes \\
\cline{3-4}
& & Input (ZH) & \small API: <knowledge> null <endofknowledge> <state> ( attraction ) consumption  " \chin{中等} " , type  " \chin{商业中心} " <endofstate> <history> USER: \chin{你好,我朋友要来苏州找我玩,我想带他找一个消费中等的商业中心逛逛,求推荐。} <endofhistory> \\
\cline{3-4}
& & Output (ZH) & \small yes \\
\cline{2-4}
& \multirow{12}{*}{DA} & \multirow{2}{*}{Input (EN)} &  \small DA: <knowledge> ( attraction ) address " Guanqian Street, Gusu District, Suzhou City. " , area " Gusu District " , available\_options " 4 " , consumption " moderate " , metro\_station " true " , name " Guanqian Street " , opening\_hours " all day " , phone\_number " N/A " , score " 4.3 " , the\_most\_suitable\_people " friends " , ticket\_price " free " , type " commercial center " <endofknowledge> <state> ( attraction ) consumption  " mid " , type  " commercial center " <endofstate> <history> USER: Hi, my friend is coming to Suzhou to visit me, I want to take him to a commercial center in the mid-price range. Do you have anything to recommend? <endofhistory> \\
\cline{3-4}
& & Output (EN) & \small ( attraction ) recommend name  " Guanqian Street " \\
\cline{3-4}
& & Input (ZH) & \small DA: <knowledge> ( attraction ) address " \chin{苏州市姑苏区观前街} " , area " \chin{姑苏区} " , available\_options " 4 " , consumption " \chin{中等} " , metro\_station " \chin{是} " , name " \chin{观前街} " , opening\_hours " \chin{全天} " , phone\_number " \chin{无} " , score " 4.3 " , the\_most\_suitable\_people " \chin{朋友出游} " , ticket\_price "\chin{免费} " , type " \chin{商业中心} " <endofknowledge> <state> ( attraction ) consumption  " \chin{中等} " , type  " \chin{商业中心} " <endofstate> <history> USER: \chin{你好,我朋友要来苏州找我玩,我想带他找一个消费中等的商业中心逛逛,求推荐。 <endofhistory>} \\
\cline{3-4}
& & Output (ZH) & \small ( attraction ) recommend name  " \chin{观前街} " \\
\cline{2-4}
& \multirow{7}{*}{RG} & \multirow{2}{*}{Input (EN)} &  \small RG: <actions> ( attraction ) recommend name  " Guanqian Street " <endofactions> <history> USER: Hi, my friend is coming to Suzhou to visit me, I want to take him to a commercial center in the mid-price range. Do you have anything to recommend? <endofhistory> \\
\cline{3-4}
& & Output (EN) & \small You can go to Guanqian Street. \\
\cline{3-4}
& & Input (ZH) & \small API: <knowledge> null <endofknowledge> <state> ( attraction ) consumption  " \chin{中等} " , type  " \chin{商业中心} " <endofstate> <history> USER: \chin{你好,我朋友要来苏州找我玩,我想带他找一个消费中等的商业中心逛逛,求推荐。} <endofhistory> \\
\cline{3-4}
& & Output (ZH) & \small \chin{去观前街吧。} \\

\specialrule{.1em}{.05em}{.05em} 
\end{tabularx}
\end{table*}

\begin{table*}
\begin{tabularx}{\linewidth}{c|c|c|X}
\specialrule{.1em}{.05em}{.05em} 
\multirow{40}{*}{\bf Turn 2} & \multirow{7}{*}{DST} & \multirow{2}{*}{Input (EN)} &  \small DST: <state> ( attraction ) consumption  " mid " , type  " commercial center " <endofstate> <history> AGENT\_ACTS: ( attraction ) recommend name  " Guanqian Street " USER: Oh yeah, why didn't I think of that? When is it open? <endofhistory> \\
\cline{3-4}
& & Output (EN) & \small ( attraction ) consumption  " mid " , name  " Guanqian Street " , the\_most\_suitable\_people  " friend " , type  " commercial center " \\
\cline{3-4}
& & Input (ZH) & \small DST: <state> ( attraction ) consumption  " \chin{中等} " , type  " \chin{商业中心} " <endofstate> <history> AGENT\_ACTS: ( attraction ) recommend name  " \chin{观前街} " USER: \chin{对呀,我怎么没想起来呢,什么时间开放呀?} <endofhistory> \\
\cline{3-4}
& & Output (ZH) & \small ( attraction ) consumption  " \chin{中等} " , name  " \chin{观前街} " , type  " \chin{商业中心} " \\
\cline{2-4}
& \multirow{10}{*}{API} & \multirow{2}{*}{Input (EN)} &  \small API: <knowledge> ( attraction ) address " \chin{苏州市姑苏区观前街} " , area " \chin{姑苏区} " , available\_options " 4 " , consumption " \chin{中等} " , metro\_station " \chin{是} " , name " \chin{观前街} " , opening\_hours " \chin{全天} " , phone\_number " \chin{无} " , score " 4.3 " , the\_most\_suitable\_people " \chin{朋友出游} " , ticket\_price " \chin{免费} " , type " \chin{商业中心} " <endofknowledge> <state> ( attraction ) consumption  " \chin{中等} " , name  " \chin{观前街} " , type  " \chin{商业中心} " <endofstate> <history> AGENT\_ACTS: ( attraction ) recommend name  " \chin{观前街} " USER: \chin{对呀,我怎么没想起来呢,什么时间开放呀?} <endofhistory> \\
\cline{3-4}
& & Output (EN) & \small yes \\
\cline{3-4}
& & Input (ZH) & \small API: <knowledge> null <endofknowledge> <state> ( attraction ) consumption  " \chin{中等} " , type  " \chin{商业中心} " <endofstate> <history> USER: \chin{你好,我朋友要来苏州找我玩,我想带他找一个消费中等的商业中心逛逛,求推荐。} <endofhistory> \\
\cline{3-4}
& & Output (ZH) & \small yes \\
\cline{2-4}
& \multirow{15}{*}{DA} & \multirow{2}{*}{Input (EN)} &  \small DA: <knowledge> ( attraction ) address " Guanqian Street, Gusu District, Suzhou City. " , area " Gusu District " , available\_options " 1 " , consumption " moderate " , features " You can try food from time-honored Suzhou brands, such as Songhelou Restaurant, Huang Tianyuan, and visit Xuanmiao Temple, the place that gave the street its name. " , metro\_station " true " , name " Guanqian Street " , opening\_hours " all day " , phone\_number " N/A " , score " 4.3 " , the\_most\_suitable\_people " friends " , ticket\_price " free " , type " commercial center " <endofknowledge> <state> ( attraction ) consumption equal\_to " mid " , name equal\_to " Guanqian Street " , the\_most\_suitable\_people equal\_to " friend " , type equal\_to " commercial center " <endofstate> <history> AGENT\_ACTS: ( attraction ) recommend name equal\_to " Guanqian Street " USER: Oh yeah, why didn't I think of that? When is it open? <endofhistory> \\
\cline{3-4}
& & Output (EN) & \small ( attraction ) inform opening\_hours  " all day " \\
\cline{3-4}
& & Input (ZH) & \small DA: <knowledge> ( attraction ) address " \chin{苏州市姑苏区观前街} " , area " \chin{姑苏区} " , available\_options " 4 " , consumption " \chin{中等} " , metro\_station " \chin{是} " , name " \chin{观前街} " , opening\_hours " \chin{全天} " , phone\_number " \chin{无} " , score " 4.3 " , the\_most\_suitable\_people " \chin{朋友出游} " , ticket\_price "\chin{免费} " , type " \chin{商业中心} " <endofknowledge> <state> ( attraction ) consumption  " \chin{中等} " , type  " \chin{商业中心} " <endofstate> <history> USER: \chin{你好,我朋友要来苏州找我玩,我想带他找一个消费中等的商业中心逛逛,求推荐。 <endofhistory>} \\
\cline{3-4}
& & Output (ZH) & \small ( attraction ) inform opening\_hours  " \chin{全天} " \\
\cline{2-4}
& \multirow{5}{*}{RG} & \multirow{2}{*}{Input (EN)} &  \small RG: <actions> ( attraction ) inform opening\_hours  " all day " <endofactions> <history> USER: Oh yeah, why didn't I think of that? When is it open? <endofhistory> \\
\cline{3-4}
& & Output (EN) & \small It's open all day. \\
\cline{3-4}
& & Input (ZH) & \small RG: <actions> ( attraction ) inform opening\_hours  " \chin{全天} " <endofactions> <history> USER: \chin{对呀,我怎么没想起来呢,什么时间开放呀?} <endofhistory> \\
\cline{3-4}
& & Output (ZH) & \small \chin{全天开放哟。} \\

\specialrule{.1em}{.05em}{.05em} 
\end{tabularx}
\caption{An example from X-RiSAWOZ validation set in Chinese and English. For brevity, only the first 2 turns are shown.}
\label{tab:examples-dataset}
\end{table*}

\subsection{Example of the Checking Process}
\begin{figure*}
    \centering
    \includegraphics[width=\linewidth]{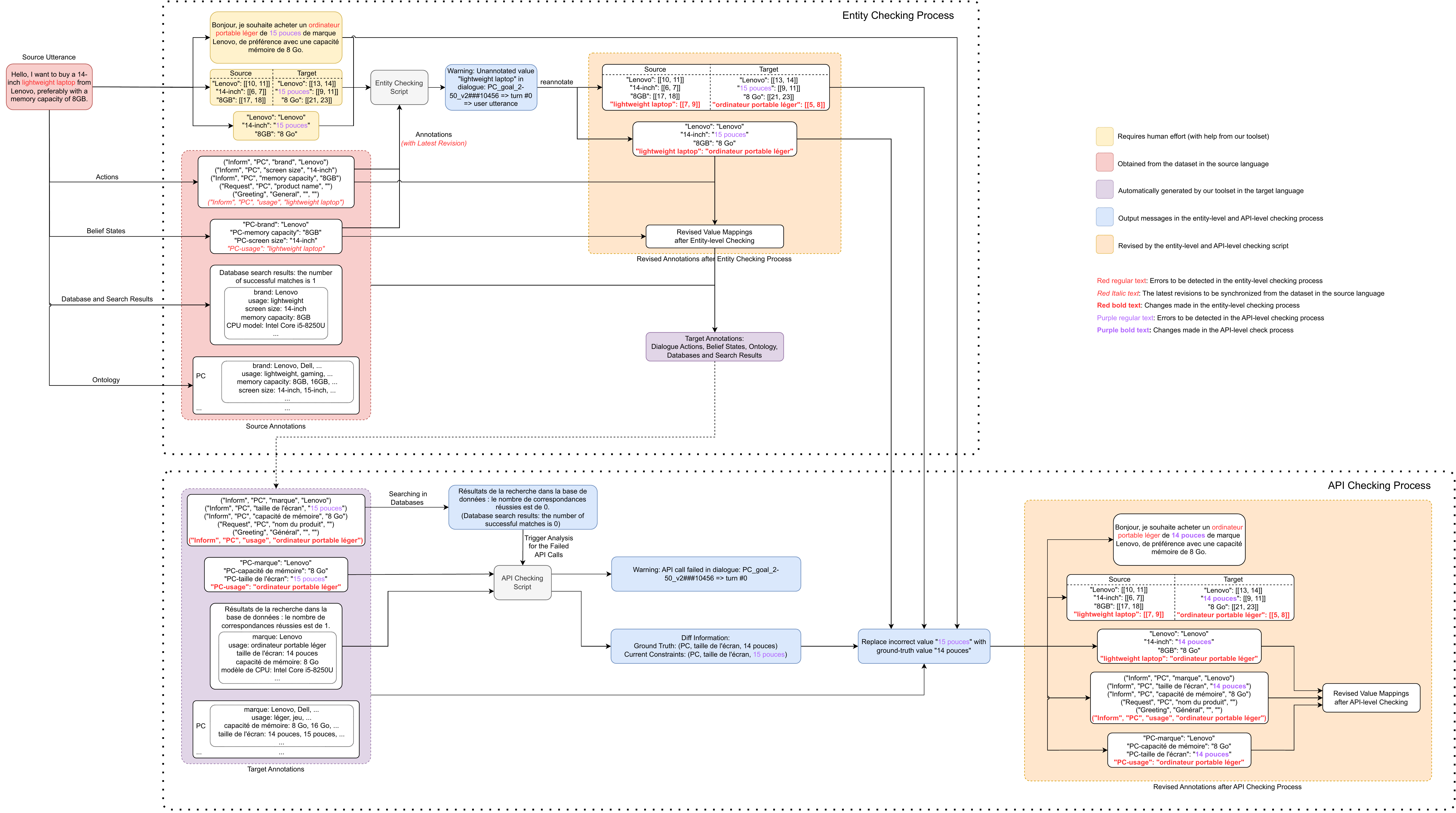}
    \caption{The entity and API checking process of X-RiSAWOZ from English to French.}
    \label{fig:dataset-checking-process}
\end{figure*}

Figure \ref{fig:dataset-checking-process} shows an example of our checking process described in Section \ref{dataset} during the translation from English to French.

\begin{figure*}
    \centering
    \includegraphics[width=\textwidth]{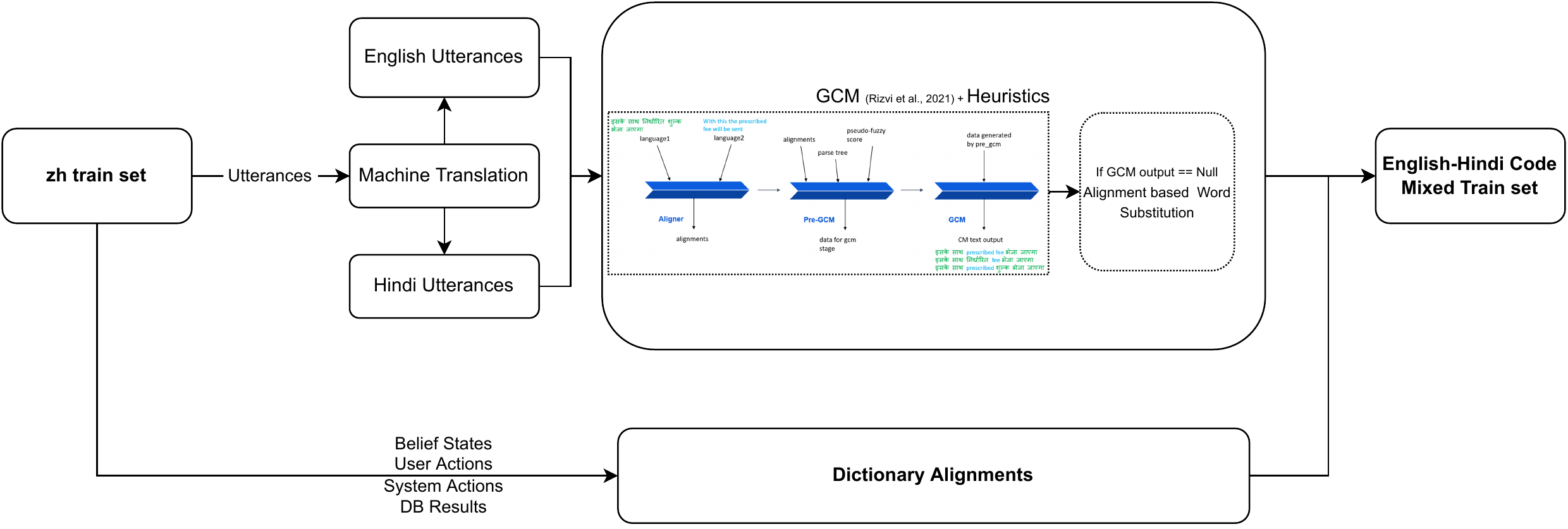}
    \caption{English-Hindi code-mixed train set is generated using a pipeline that combines GCM~\citep{rizvi-etal-2021-gcm}, and word alignment to generate code-mixed utterances. Entities in Belief states, user and system actions are substituted using dictionary alignment.}
    \label{fig:enhi-trainset}
\end{figure*}

\end{document}